\documentclass[10pt,twocolumn,letterpaper]{article}

\usepackage[pagenumbers]{cvpr} 

\usepackage[accsupp]{axessibility}  

\usepackage{graphicx}
\usepackage{amsmath}
\usepackage{amssymb}
\usepackage{booktabs}

\usepackage{diagbox, eqparbox}
\usepackage{tabularx}
\usepackage{tabu, multirow}
\usepackage{soul}
\usepackage{makecell}
\usepackage[inline]{enumitem}
\usepackage[symbol]{footmisc}

\usepackage{caption}
\usepackage{threeparttable}
\usepackage{bm}

\newcommand{\mpiitarget}{MPII-NV}

\newif\ifdraft
\drafttrue
\draftfalse
 
\usepackage{xcolor}
\usepackage[normalem]{ulem}

\ifdraft
    \newcommand{\key}[1]{\textcolor{red}{{Key: #1}}}
    
    \newcommand{\todo}[1]{\textcolor{red}{\textbf{TODO: #1}}}
    \newcommand{\sugano}[1]{\textcolor{orange}{{[Sugano: #1]}}}
    \newcommand{\jqin}[1]{\textcolor{magenta}{{[Jiawei: #1]}}}
    \newcommand{\tshimo}[1]{\textcolor{cyan}{{[Shimoyama: #1]}}}
    \newcommand{\review}[1]{\textcolor{blue}{{[Reviewer: #1]}}}
    \newcommand{\del}[1]{\textcolor{red}{\sout{#1}}}
    
\else
    \newcommand{\key}[1]{}
    
    \newcommand{\todo}[1]{}
    \newcommand{\sugano}[1]{}
    \newcommand{\jqin}[1]{}
    \newcommand{\tshimo}[1]{}
    \newcommand{\review}[1]{}
    \newcommand{\del}[1]{}
    
\fi

\usepackage[pagebackref,breaklinks,colorlinks]{hyperref}

\usepackage[capitalize]{cleveref}
\crefname{section}{Sec.}{Secs.}
\Crefname{section}{Section}{Sections}
\Crefname{table}{Table}{Tables}
\crefname{table}{Tab.}{Tabs.}

\def\cvprPaperID{2} 
\def\confName{CVPR}
\def\confYear{2022}

\begin{document}


\title{Learning-by-Novel-View-Synthesis for \\ Full-Face Appearance-Based 3D Gaze Estimation}

\author{Jiawei Qin, Takuru Shimoyama, Yusuke Sugano\\
Institute of Industrial Science, The University of Tokyo\\
{\tt\small \{jqin, tshimo, sugano\}@iis.u-tokyo.ac.jp}
}
\maketitle

\begin{abstract}
Despite recent advances in appearance-based gaze estimation techniques, the need for training data that covers the target head pose and gaze distribution remains a crucial challenge for practical deployment.
This work examines a novel approach for synthesizing gaze estimation training data based on monocular 3D face reconstruction.
Unlike prior works using multi-view reconstruction, photo-realistic CG models, or generative neural networks, our approach can manipulate and extend the head pose range of existing training data without any additional requirements.
We introduce a projective matching procedure to align the reconstructed 3D facial mesh with the camera coordinate system and synthesize face images with accurate gaze labels.
We also propose a mask-guided gaze estimation model and data augmentation strategies to further improve the estimation accuracy by taking advantage of synthetic training data.
Experiments using multiple public datasets show that our approach significantly improves the estimation performance on challenging cross-dataset settings with non-overlapping gaze distributions.

\end{abstract}

\section{Introduction}
Gaze estimation has been considered an important research topic in the computer vision community with many applications.
Vision-based techniques have the potential to bring the ability to estimate gaze to arbitrary cameras. 
However, despite recent advances in machine learning-based approaches~\cite{Park2018ECCV, Sugano_2014_CVPR, zhang2017s, Zhang2020ETHXGaze, zhang15_cvpr}, it is still challenging to accurately predict gaze directions under extreme head poses and diverse lighting conditions.

One of the fundamental difficulties is the requirement of an appropriate training dataset.
Many efforts have been made to create diverse in-the-wild gaze datasets~\cite{zhang15_cvpr, 7780608, Fischer_2018_ECCV, gaze360_2019}. 
However, it is not a trivial task to construct a dataset covering all crucial factors including head pose and gaze distribution, illumination environment, background appearance, demographic diversity, imaging properties, and accurate gaze labels.

As actively studied in other computer vision tasks~\cite{Su_2015_ICCV, Gupta_2016_CVPR, Sankaranarayanan_2018_CVPR, 8575297, Richter_2016_ECCV, Wood_2021_ICCV}, one potential approach to obtain targeted training data is the use of synthetic images. 
In the context of appearance-based gaze estimation, accurate ground-truth 3D gaze direction is required when synthesizing images.
Previous approaches use either multi-view 3D reconstructed data~\cite{5226635, Sugano_2014_CVPR} or hand-crafted eyeball models~\cite{7410785} to synthesize eye images for appearance-based gaze estimation.
However, it is still challenging to capture 3D reconstruction data under various illumination conditions.
While hand-crafted computer graphics models can potentially address this limitation, there is also a vast domain gap between real and synthetic images~\cite{Shrivastava_2017_CVPR}.
These limitations become more prominent for full-face gaze estimation~\cite{zhang2017s, 7780608, Zhang2020ETHXGaze}.
Although generative neural rendering models are one of the promising approaches to generate full-face images while controlling gaze directions~\cite{Zheng2020NeurIPS}, it is not easy to ensure that their labels are accurate enough to be used as training data.

\begin{figure}[t]
\begin{center}
  \includegraphics[width=0.75\linewidth]{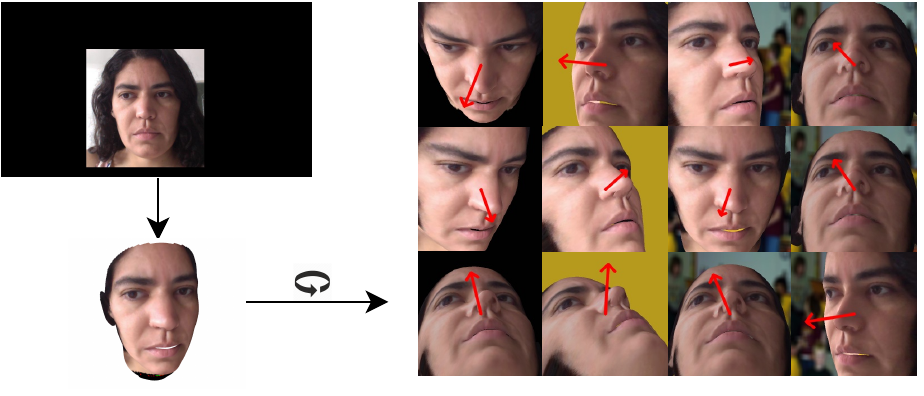}
\end{center}
  \caption{We propose a learning-by-synthesis appearance-based gaze estimation approach based on single-image 3D face reconstruction. Our projective matching procedure aligns the reconstructed face with the ground-truth gaze position for generating precise training data.}
\label{fig:teaser}
\end{figure}

This work proposes an alternative approach to learning-by-synthesis full-face appearance-based gaze estimation via single-image 3D face reconstruction.
As illustrated in Fig.~\ref{fig:teaser}, we reconstruct 3D facial shapes from existing gaze datasets and synthesize novel views by rotating the reconstructed faces.
However, since most of the single-image 3D face reconstruction methods do not provide physical 3D shapes in the camera coordinate system, it brings another challenge of preserving accurate gaze labels under novel views.
To address this issue, we introduce a projective matching procedure to ensure that the reconstructed 3D facial surface is associated with the original camera coordinate system and the ground-truth gaze target position.

In addition, we propose a novel mask-guided gaze estimation model.
We take full advantage of the data synthesis by obtaining a facial region mask during rendering process and using it as an additional supervision.
We also propose rendering images with lighting and background augmentation to enhance the diversity of the image appearance.
We evaluate how the proposed approach can cover unseen head poses and gaze directions by the data extrapolation task.
By combining our synthetic data and mask-guided estimation model, we show that our approach can outperform the gaze estimation results of other state-of-the-art synthetic- and real-image training datasets.

The contributions of this work are threefold.
\begin{enumerate*} [label=(\roman*)]
\item We propose a novel approach for creating training data for appearance-based gaze estimation through monocular 3D face reconstruction.
To our knowledge, this is the first work to prove that single-image face reconstruction outputs can be used to train full-face appearance-based gaze estimation models.
\item We propose a novel mask-guided soft-attention model for gaze estimation. 
Together with data augmentation, our gaze estimation method fully utilizes the nature of synthetic training. 
\item Through experiments, we verify that our approach can successfully extend the gaze range of the source dataset, which provides better model performance than other baseline training datasets using real and synthetic images.
\end{enumerate*}
\section{Related Work}

\begin{figure*}[ht]
\begin{center}
  \includegraphics[width=0.75\linewidth]{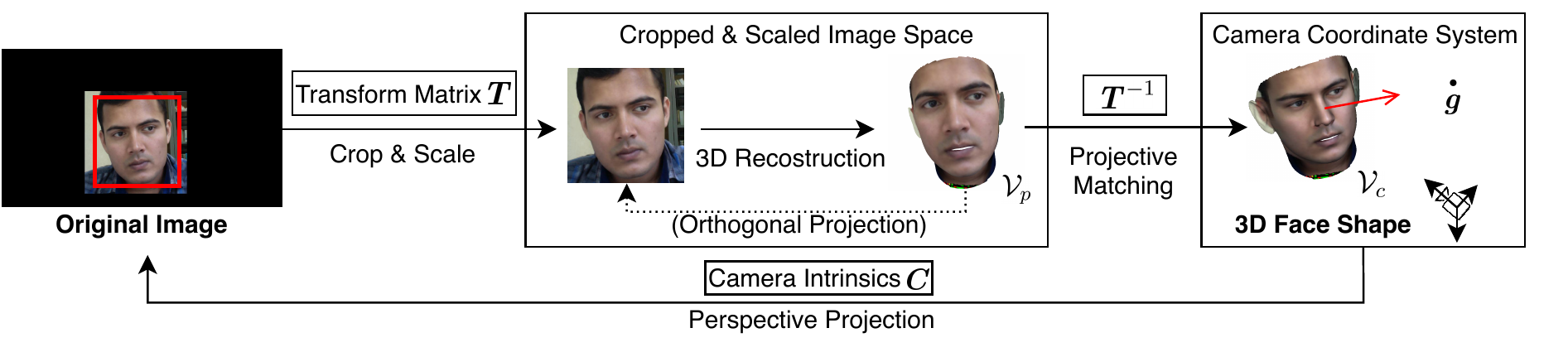}
\end{center}
  \caption{Overview of our data synthesis pipeline. We assume that 3D face reconstruction methods generate facial meshes under an orthogonal projection model, and we convert the mesh via the proposed projective matching to align with the ground-truth gaze position in the input camera coordinate system. 
  }
\label{fig:overview}
\end{figure*}

Traditional model-based gaze estimation methods use 3D eyeball models with geometric features to infer 3D gaze directions~\cite{1634506,hansen2009eye}. 
On the other hand, appearance-based gaze estimation methods directly map the image to gaze direction~\cite{Tan2002AppearancebasedEG}. 
Methods in this category have fewer hardware restrictions and are more suitable for in-the-wild settings. 

Most appearance-based methods take eye-only images as input~\cite{4449972, Tan2002AppearancebasedEG, Sugano_2014_CVPR, zhang15_cvpr, xiong2019mixed, yu2019improving, yu2020unsupervised}, and there have also been some attempts to explore two-eye combination inputs~\cite{9050633, Cheng_2018_ECCV, Park2019ICCV, guo2020domain}. 
In contrast, some prior work demonstrated that full-face input can improve the robustness and accuracy of appearance-based gaze estimation~\cite{7780608, zhang2017s, Zhang2020LearningbasedRS, cheng2020coarse}.
While this work also focuses on the full-face appearance-based gaze estimation task, we explore the potential of using single-image 3D face reconstruction to synthesize full-face training data for the first time.


\textbf{Gaze Estimation Datasets}.
Although some datasets were collected using mobile devices during in-the-wild daily-life situations under diverse illumination conditions, they often suffer from limited ranges of gaze and head pose~\cite{zhang19_pami,zhang2017s, zhang15_cvpr, huang2017tabletgaze, 7780608}. 
Some datasets reached higher variety in head pose and gaze by using more complex recording setups, but the environment and illumination are always limited to controlled conditions~\cite{CAVE_0324, Fischer_2018_ECCV, FunesMora_ETRA_2014}.

Recent datasets have been collected with further extended diversity in head pose ranges and environment conditions~\cite{Zhang2020ETHXGaze, gaze360_2019}.
However, a significant effort is still required to acquire training datasets that meet the requirement for head pose and appearance variations in the deployment environment.
This work aims to address this issue by providing a method for extending the head pose ranges of source datasets as well as augmenting the environment diversity.

\textbf{Learning-by-Synthesis for Gaze Estimation}.
To address the limitations of real-world data collection, there have been some efforts on creating synthetic training data for appearance-based gaze estimation using multi-view stereo reconstruction~\cite{Sugano_2014_CVPR} or hand-crafted photo-realistic computer graphics models~\cite{egp.20161054, wood2016_etra}.
However, the multi-view setup has the fundamental limitation that the environment is fixed to the laboratory conditions~\cite{Sugano_2014_CVPR}, and the domain gap between real and purely synthetic images is not negligible~\cite{egp.20161054, wood2016_etra}. 
Zheng~\etal proposed a neural network for redirecting gaze and head pose, which can be also used to generate synthetic training data~\cite{Zheng2020NeurIPS}. 
However, such neural rendering models cannot guarantee that the facial appearance exactly matches the target gaze label.
In this work, we take yet another approach based on single-image 3D face reconstruction for accurate data synthesis.

\textbf{Domain Adaptation for Gaze Estimation}.
When using synthetic data, the domain gap between synthetic and real images can be a critical issue. 
However, in the context of appearance-based gaze estimation, there have been few studies dealing with such an unsupervised, cross-environment domain adaptation task.
Fundamentally speaking, there have been few research examples of domain adaptation for regression tasks~\cite{9009467, takahashi2020partially}. 
Shrivastava~\etal~\cite{Shrivastava_2017_CVPR} proposed SimGAN, an unsupervised domain adaptation approach that refines synthetic eye images to be visually similar to real images.
However, their method was designed for eye images, and its effectiveness has never been validated on full-face gaze estimation.
Liu~\etal~\cite{liu2021PnP_GA} recently proposed an unsupervised domain adaptation framework based on collaborative learning. 
Although their work addresses the full-face gaze estimation task, its effectiveness on synthetic source data has not been evaluated.
In contrast to these methods taking domain adaptation approaches, we propose a method that addresses the domain gap by fully utilizing the characteristics of the synthetic training data.


\textbf{3D Face Reconstruction}.
Monocular 3D face reconstruction techniques have also made significant progress in recent years~\cite{Zollhfer2018StateOT}. 
While reconstructed 3D faces have also been used to augment face recognition training data~\cite{1301639, zhou2020rotate, 7961797}, no prior work explored its usage in full-face appearance-based gaze estimation.
Methods based on 3D morphable models~\cite{deng2019accurate, Tran_2017_CVPR} usually approximate facial textures via the appearance basis~\cite{10.1145/311535.311556, bfm09,FLAME:SiggraphAsia2017, ploumpis2020towards}, and therefore the appearances of the eye region can be distorted. 
To preserve accurate gaze labels after reconstruction, this work utilizes 3D face reconstruction methods that sample texture directly from the input image~\cite{6412675, Zhu_2016_CVPR, bulat2017far, guo2020towards, Yao2021DECA, 3ddfa_cleardusk, zhu2017face}. 
In addition, since many prior works rely on orthogonal or weak perspective projection models, we discuss how to precisely align the reconstruction results with the source camera coordinate system.
\section{View Synthesis via 3D Face Reconstruction}

Given an ordinary single-view gaze dataset and 3D face reconstruction results, our goal is to synthesize face images under unseen head poses while preserving accurate gaze direction annotations.

\subsection{Overview}

Fig.~\ref{fig:overview} shows the overview of our data synthesis pipeline.
We assume that the source gaze dataset consists of 1) face images, 2) the projection matrix (intrinsic parameters) $\bm{C}$ of the camera, and 3) the 3D gaze target position $\bm{g} \in \mathbb{R}^3$ in the camera coordinate system.
Most of the existing gaze datasets contain 3D gaze position annotations~\cite{zhang15_cvpr,7780608, Zhang2020ETHXGaze}, and yaw-pitch annotations can also be converted assuming a distance to the dummy target.
State-of-the-art learning-based 3D face reconstruction methods usually take a cropped face patch as input and output a 3D facial mesh, which is associated with the input image in an orthographic projection way.
Without loss of generality, we assume that the face reconstruction method takes a face bounding box defined with center $(c_x, c_y)$, width $w_b$, and height $h_b$ in pixels and then resized to a fixed input size by factor $(s_x, s_y)$.
The reconstructed facial mesh is defined as a group of $N$ vertices $\mathcal{V}_{p}=\{\bm{v}_p^{(i)}\}_{i=0}^N$.
Each vertex is represented as $\bm{v}_p^{(i)} = [u^{(i)},v^{(i)},d^{(i)}]^\top$ in the right-handed coordinate system, where $u$ and $v$ directly correspond to the pixel locations in the input face patch and $d$ is the distance to the $u$-$v$ plane in the same pixel unit. 
Many recent works use this representation~\cite{3ddfa_cleardusk, feng2018prn, Zhu_2016_CVPR, Jourabloo_2015_ICCV, bulat2017far}, and we can convert arbitrary 3D representation to this way by projecting the reconstructed 3D face onto the input face patch.

Our goal is to convert the vertices of the reconstructed 3D face $\mathcal{V}_{p}$ to another 3D representation $\mathcal{V}_{c}=\{\bm{v}_c^{(i)}\}_{i=0}^N$ where each vertex $\bm{v}_c^{(i)} = [x^{(i)},y^{(i)},z^{(i)}]^\top$ is in the original camera coordinate system so that it can be associated with the gaze annotation $\bm{g}$.
In this way, the gaze target location can be also represented in the facial mesh coordinate system, and we can render the facial mesh under arbitrary head or camera poses together with the ground-truth gaze direction information.

\subsection{Projective Matching}\label{projective-matching}

Since $u$ and $v$ of the each reconstructed vertex $\bm{v}_{p}$ are assumed to be aligned with the face patch coordinate system, $\bm{v}_{c}$ must be on the back-projected ray as 
\begin{equation}\label{eq:invproj}
\bm{v}_{c} =
\lambda \frac{\bm{C}^{-1} \bm{p}_o}{||\bm{C}^{-1} \bm{p}_o||} =
\lambda \frac{\bm{C}^{-1} \bm{T}^{-1} \bm{p}}{||\bm{C}^{-1} \bm{T}^{-1} \bm{p}||},
\end{equation}
where $\bm{p}_o = [u_o, v_o, 1]^{\top}$ and $\bm{p} = [u, v, 1]^{\top}$ indicates the pixel locations in the original image and the face patch in the homogeneous coordinate system, respectively, and
\begin{equation}
\bm{T} = \begin{bmatrix} s_x & 0  & - s_x (c_x - \frac{w}{2}) \\0 & s_y & -s_y (c_y - \frac{h}{2}) \\0 & 0 &1  \end{bmatrix}
\end{equation}
represents the cropping and resizing operation to create the face patch, \ie, $\bm{p} = \bm{T}\bm{p}_o$.
The scalar $\lambda$ indicates scaling along the back-projection ray and physically means the distance between the camera origin and $\bm{v}_{c}$.

\begin{figure}[t]
\begin{center}
  \includegraphics[width=0.7\linewidth]{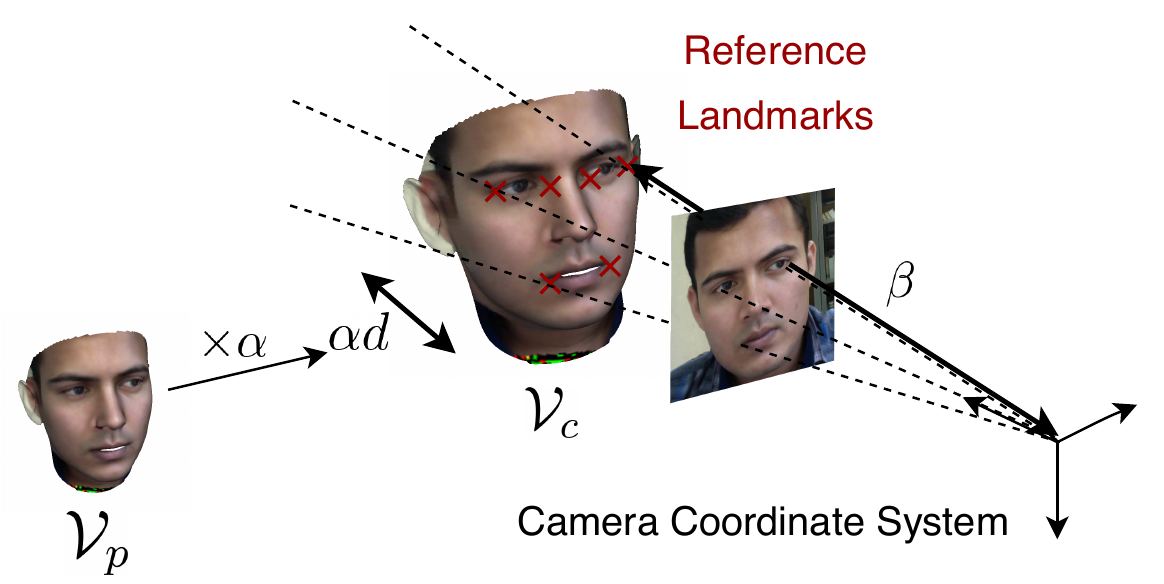}
\end{center}
  \caption{Determining the location of $\mathcal{V}_{c}$ via parameters $\alpha$ and $\beta$. $\alpha$ indicates a scaling factor from the pixel to physical (\eg, millimeter) unit, and $\beta$ is the bias term to align $\alpha d$ to the camera coordinate system.}
\label{fig:alpha-beta}
\end{figure}

Since Eq.~(\ref{eq:invproj}) does not explain anything about $d$, our task can be understood as finding $\lambda$ which also maintains the relationship between $u$, $v$, and $d$.
Therefore, as illustrated in Fig.~\ref{fig:alpha-beta}, we propose to define $\lambda$ as a function of $d$ as $\lambda = \alpha d + \beta$.
$\alpha$ indicates a scaling factor from the pixel to physical (\eg, millimeter) unit, and $\beta$ is the bias term to align $\alpha d$ with the camera coordinate system.
Please note that $\alpha$ and $\beta$ are constant parameters determined for each input image and applied to all of the vertices from the same image.

We first fix $\alpha$ based on the distance between two eye centers (midpoints of two eye corner landmarks) in comparison with a physical reference 3D face model.
3D face reconstruction methods usually require facial landmark detection as a pre-processing step, and we can naturally assume that we know the corresponding vertices in $\mathcal{V}_{p}$ to the eye corner landmarks.
We use a 3D face model with 68 landmarks (taken from the OpenFace library~\cite{8373812}) as our reference.
We set $\alpha = l_{r}/l_{p}$, where $l_{p}$ and $l_{r}$ are the eye-center distances in $\mathcal{V}_{p}$ and in the reference model, respectively.

We then determine $\beta$ by aligning the reference landmark depth in the camera coordinate system. 
In this work, we use the face center as a reference, which is defined as the centroid of the eyes and the mouth corner landmarks, following previous works on full-face gaze estimation~\cite{zhang2017s, zhang18_etra}, and we use the same face center as the origin of the gaze vector through the data normalization and the gaze estimation task.

We approximate $\beta$ as the distance between the ground-truth 3D reference location and the scaled/reconstructed location as $\beta = ||\bar{\bm{v}}|| - \alpha \bar{d}$.
$\bar{d}$ is the reconstructed depth values computed as the mean of six landmark vertices corresponding to the eye and mouth corner obtained in a similar way as when computing $\alpha$.
$\bar{\bm{v}}$ is the centroid of the 3D locations of the same six landmarks in the camera coordinate system, which are obtained by minimizing the projection error of the reference 3D model to the 2D landmark locations using the Perspective-n-Point (PnP) algorithm~\cite{10.1145/358669.358692}.

\subsection{Training Data Synthesis}\label{rotate-and-render}

\begin{figure*}[t]
\begin{center}
  \includegraphics[width=0.78\linewidth]{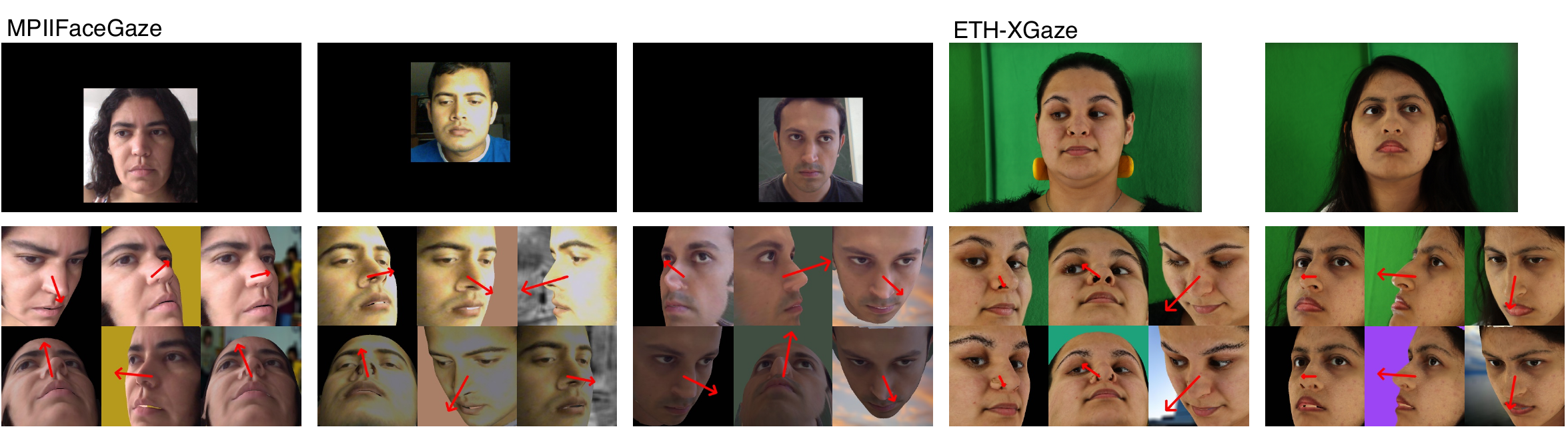}
\end{center}
  \caption{Examples of the synthesized images. The first row shows the source images from MPIIFaceGaze~\cite{zhang2017s} and ETH-XGaze~\cite{Zhang2020ETHXGaze}. The second and third rows show synthesized images for MPIIFaceGaze, and pairs of real and synthesized images for ETH-XGaze. For MPIIFaceGaze, the second and third rows show synthesized images in full-light and weak-light. For ETH-XGaze, the second row shows the real images from the dataset, and the third row shows our synthetic images with the same head poses as the second row. For each synthetic example, the three columns show the black, color, and scene image background in turn, and the red arrows indicate gaze direction vectors.} 
\label{fig:render-example}
\end{figure*}

Once we obtain the 3D face mesh $\mathcal{V}_{c}$ in the original camera coordinate system, we can render it under arbitrary head poses with the ground-truth gaze vector.

If our goal is to render a face image in a new camera coordinate system which is defined with extrinsic parameters $\bm{R}_e, \bm{t}_e$, the vertex $\bm{v}_c$ and gaze target position $\bm{g}$ are both projected to the new coordinate system as $\bm{R}_e \bm{v}_{c} + \bm{t}_e$ and $\bm{R}_e \bm{g} + \bm{t}_e$ in the same manner.
Similarly, if the goal is to render a face image with a target head pose\footnote{Head pose is defined as the rotation and translation from the face coordinate system to the camera coordinate system.} $\bm{R}_t, \bm{t}_t$ in a new camera coordinate system given the source head pose $\bm{R}_s, \bm{t}_s$, we can transform the vertices and gaze position as $ \bm{R}_t (\bm{R}_s)^{-1} (\bm{v}_c -\bm{t}_s) + \bm{t}_t$.

In this work, we further augment the images in terms of lighting conditions and background appearances by virtue of the flexible synthetic rendering.
We set background to random color or random scene images.
Although most of the 3D face reconstruction methods do not reconstruct lighting and albedo, we maximize the diversity of rendered images by controlling the global illumination.
We randomly reduce the ambient light intensity to render darker weak light images.
Fig.~\ref{fig:render-example} shows examples of the synthesized images using MPIIFaceGaze~\cite{zhang2017s} and ETH-XGaze~\cite{Zhang2020ETHXGaze}.


\subsection{Rendering Details}

In the experiments, we applied 3DDFA~\cite{3ddfa_cleardusk} to reconstruct 3D faces from the source dataset.
After projective matching, we rendered new images using the PyTorch3D library~\cite{ravi2020pytorch3d}.
We set the background to be a random RGB value or scene image by modifying the blending setting.
In the PyTorch3D renderer, the ambient color $[r,g,b]$ 
represents the ambient light intensity, ranging from $0$ to $1$, in which $1$ is the default value for full lighting.
For weak-light images, we set them to be a random value between $0.25$ and $0.75$.
Overall, among all generated images, the ratio of black, random color, and random scene are set to 1:1:3, and half of them are weak lighting.
Random scene images are taken from the Places365 dataset~\cite{zhou2017places} and we apply blurring to them before rendering faces.

\section{Mask-Guided Gaze Estimation}
While the data synthesis process described above can render realistic face regions with accurate gaze labels, there still remains a huge synthetic-real appearance gap.
We cannot fully ignore the influence of background and non-face (\eg, hair and clothes) regions in the full face estimation task. 
Synthesizing invisible face regions of the original image is difficult even with the state-of-the-art face reconstruction methods.
In this section, we describe our mask-guided gaze estimation model that addresses the domain gap issue by additional supervision obtained from data synthesis.

\subsection{Network Architecture}

\begin{figure}[t]
\begin{center}
  \includegraphics[width=1.0\linewidth]{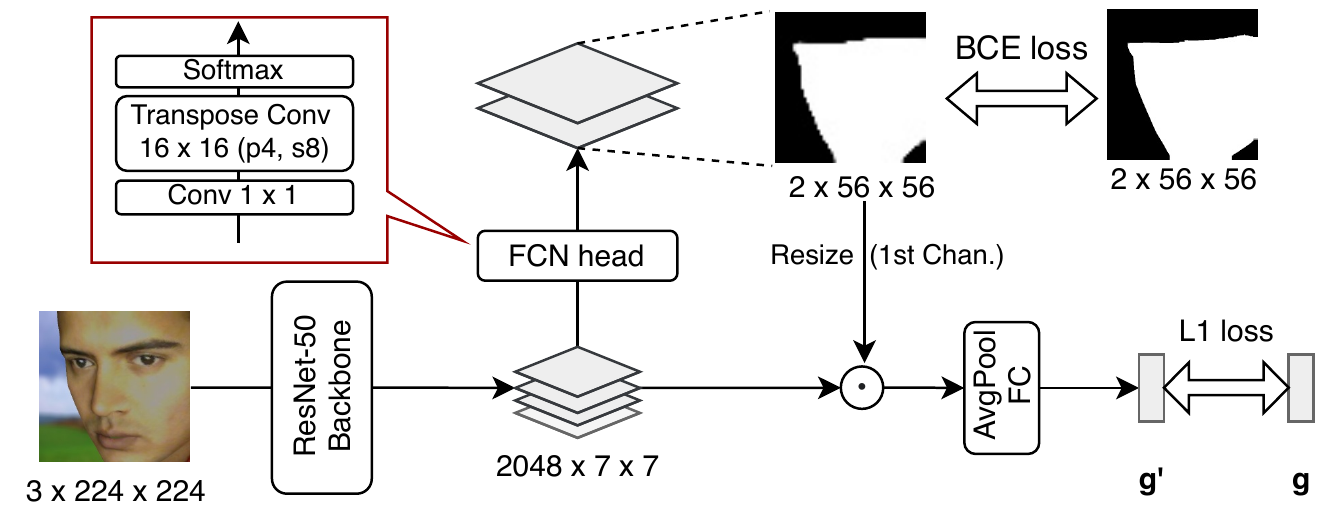}
\end{center}
  \caption{ Architecture of the proposed mask-guided gaze estimation network with an extra segmentation branch whose output segmentation mask serves as a soft attention.}
\label{fig:mask-guided-model}
\end{figure}

Fig.~\ref{fig:mask-guided-model} shows the overview of the proposed mask-guided gaze estimation network.
When synthesizing the training data, we propose to generate binary masks representing the reliable regions of the reconstructed facial mesh, \eg, the frontal face regions visible in the source image.
In addition to the base gaze estimation network, our proposed network has an extra fully-convolutional branch~\cite{Long_2015_CVPR} after the feature extractor to predict segmentation masks corresponding to such synthesized binary masks.
The output segmentation mask is then applied to the feature map, serving as a soft attention~\cite{10.5555/3045118.3045336} to enhance informative feature regions.

The network is trained in a multitask manner by combining two loss functions as $\mathcal{L} =  \mathcal{L}_{\textrm{gaze}} + \gamma \mathcal{L}_{\textrm{mask}}$,
where $\mathcal{L}_{\textrm{gaze}}$ and $\mathcal{L}_{\textrm{mask}}$ correspond to loss terms evaluating the gaze direction and the segmentation mask, respectively.
Following~\cite{Zhang2020ETHXGaze}, $\mathcal{L}_{\textrm{gaze}}$ is defined as an $\ell1$ loss between the ground truth and the predicted gaze direction.
$\mathcal{L}_{\textrm{mask}}$ is defined as a binary cross-entropy loss between the ground-truth region mask and the predicted segmentation mask.


\subsection{Implementation Details}
Together with the synthetic face images, we also generated mask images to supervise the training. 
As most of the reconstruction source images are nearly frontal faces, and the reconstructed 3D surfaces are aligned with the input image, we use the 2D landmark locations to define the face region outline and filter out the 3D vertices outside the region. 
We also filter out the vertices with depth values larger than that of the jaw-landmark, and finally obtain the face-region-only vertices for rendering binary masks.

In the following experiments, we use ImageNet pre-trained ResNet-50~\cite{7780459} as the backbone network.
We predict a two-class segmentation mask using a FCN head, whose architecture is shown in Fig.~\ref{fig:mask-guided-model}.
We resize its first channel to $7\times7$ and multiply it by the original feature map element-wisely.
We compute the binary cross-entropy loss using the ground-truth face region mask with an extra bitwise inverted channel.
The loss weight $\gamma$ is set to be $0.5$.

\section{Experiments}

We conduct experimental evaluations to show the feasibility of our approach to synthesize training datasets.
We compare our method with existing real datasets and data synthesis approach in terms of gaze estimation accuracy.

\begin{figure*}[t]
\begin{center}
  \includegraphics[width=0.9\linewidth]{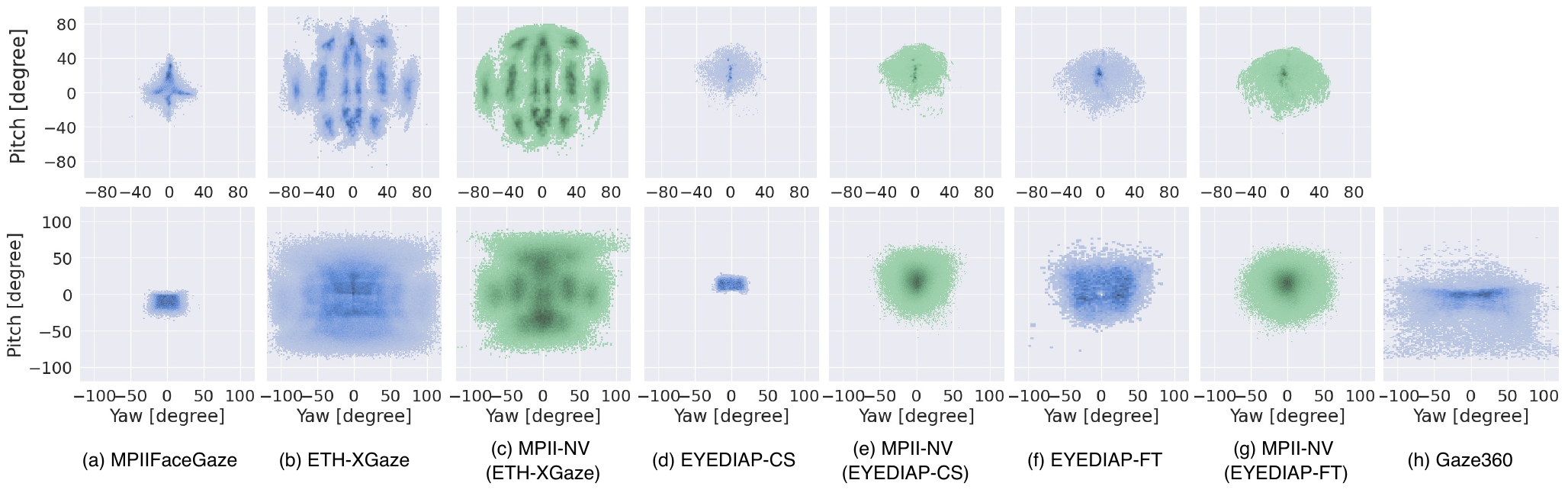}
\end{center}
  \caption{Distributions of head pose (top row) and gaze direction (bottom row). (a) source MPIIFaceGaze, (b) target ETH-XGaze, (c) synthesized dataset by extending MPIIFaceGaze for ETH-XGaze distribution, (d) target EYEDIAP (CS), (e) synthesized dataset for EYEDIAP (CS), (f) target EYEDIAP (FT), (g) synthesized dataset for EYEDIAP (FT), and (h) Gaze360 (head pose not provided).}
\label{fig:distribution}
\end{figure*}

\subsection{Experimental Settings}

\textbf{MPIIFaceGaze~\cite{zhang2017s}} consists of more than 38,000 images of 15 subjects.
Since we use this dataset only as a source for synthesis, we restricted the source images to be nearly frontal and removed reconstruction failure cases. 
To ensure subject balance for training, we randomly down-sample or up-sample to 1,500 images for each subject.
\textbf{ETH-XGaze~\cite{Zhang2020ETHXGaze}} contains more than 1 million images of 110 subjects.
For its nonpublic-label testing set, we use the public evaluation server for evaluating the accuracy.
\textbf{EYEDIAP~\cite{FunesMora_ETRA_2014}} consists of more than 4 hours of video data captured by VGA and HD cameras, using continuous screen targets or 3D floating object targets. 
We treated the screen target (CS) and floating target (FT) subsets separately and sampled one image every 5 frames from the VGA videos following the pre-processing by Park~\etal~\cite{Park2019ICCV}.
\textbf{GazeCapture~\cite{7780608}} consists of more than 2 million images crowdsourced from more than 1,300 subjects. 
We used the metadata provided by Park~\etal~\cite{Park2019ICCV} for data normalization.
\textbf{Gaze360~\cite{gaze360_2019}} consists of indoor and outdoor images of 238 subjects with a very large head pose and gaze range. 
We follow the pre-processing of Cheng~\etal~\cite{Cheng2021Survey}, which omits the cases of invisible eyes, resulting in 84,902 images. 

We apply the data normalization scheme commonly used in appearance-based gaze estimation~\cite{zhang18_etra,Zhang2020ETHXGaze}.
Unless otherwise noted, we follow the ETH-XGaze dataset~\cite{Zhang2020ETHXGaze}.
We directly render the 3D facial mesh in the normalized camera space.
We set the virtual camera's focal length to $960$ mm, and the distance from the camera origin to the face center to $300$ mm.
Face images are rendered in $448\times448$ pixels and down-scaled to $224\times224$ pixels before being fed into CNNs.
3D head pose is obtained by fitting a $6$-landmark 3D face model to the 2D landmark locations provided by the datasets, using the PnP algorithm~\cite{10.1145/358669.358692}.
We apply the rotation matrix to rotate the 3D facial mesh to a normalized target head pose. 
For some source images, there may exist a misalignment between the estimated head poses and the 3D facial mesh, which would result in an in-plane rotation after rotating the mesh.
We address this by applying an extra rotation.
Specifically, we determine the $x$, $y$, $z$-axis based on the face mesh's 3D landmarks.
Equally, the extra rotation is also multiplied on the gaze vector and the head pose to update the labels consistently.
Although this does not compensate the misalignment, it ensures the rendered face has no in-plane rotation, while keeping the gaze label correct.

As a simple baseline model against our mask-guided network, we use a gaze estimation network with the ResNet-50~\cite{7780459}.
This corresponds to the proposed network without the segmentation branch and the attention mechanism, which is evaluated as a baseline model in ETH-XGaze~\cite{Zhang2020ETHXGaze}.


\subsection{Dataset Extrapolation}

We first focus on the dataset extrapolation cases where the source MPIIFaceGaze dataset is extended to have a similar head pose distribution as the target ETH-XGaze\footnote{We used the training subset as the target head pose distribution.} and EYEDIAP datasets.
We use the head pose values obtained through the data normalization process, and each source image is reconstructed and rendered with 16 new head poses randomly chosen from the target dataset.
To avoid extreme profile faces where the eyes are fully occluded, we discarded the cases whose pitch-yaw vector norm is larger than 80 degrees.
As a result, the MPIIFaceGaze is extended to three synthetic datasets for ETH-XGaze, EYEDIAP CS, and EYEDIAP FT, respectively, all with 360,000 images.
We refer to these datasets as \mpiitarget.

We evaluate how our data synthesis approach improves performance compared to other baseline training data.
As a real image baseline, we used the Gaze360 dataset which mostly covers the target gaze range.
The head pose and gaze distributions of the source and target real datasets (blue) and the synthetic datasets (green) are shown in Fig.~\ref{fig:distribution}, together with the gaze distribution of Gaze360 (head pose is not provided).
Since we synthesized the data based on head pose distribution, it can be seen that the gaze distribution does not exactly match the target, but only roughly overlaps.

\begin{figure}[t]
\begin{center}
  \includegraphics[width=0.9\linewidth]{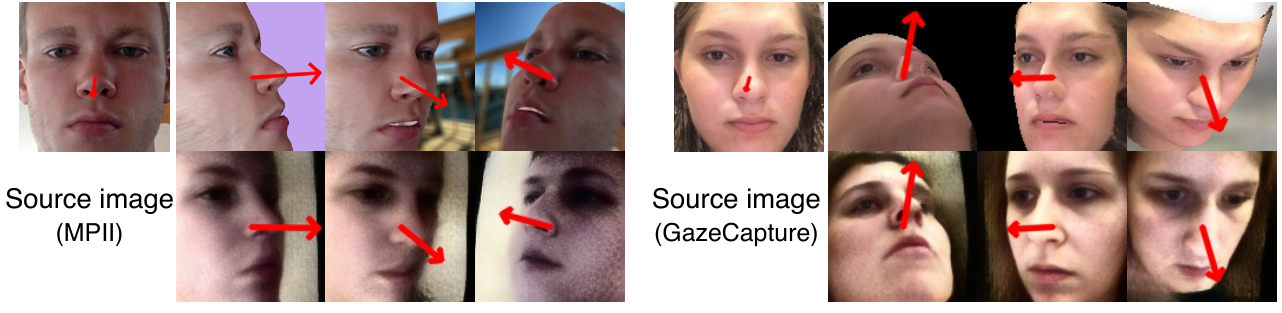}
\end{center}
  \caption{Examples of the synthesized images.
  For each source image, the first row is from our proposed method, and the second row is from ST-ED~\cite{Zheng2020NeurIPS}.}
\label{fig:sted-samples}
\end{figure}

In addition, we use ST-ED~\cite{Zheng2020NeurIPS} as a neural rendering-based synthetic baseline.
We used their pre-trained model and multiplied the same rotation matrix on the head pose and gaze embeddings, so that each image is rotated in the same manner as \mpiitarget.
The dataset is named MPII-NV-STED whose samples are shown in the bottom left of Fig.~\ref{fig:sted-samples}.
Since the pre-trained model of ST-ED can only output $128 \times 128$ images, we downscaled the test images when evaluating the model trained on ST-ED.
We also show results of \mpiitarget~downscaled to $128 \times 128$ for a fair comparison.

\begin{table}[t]
\begin{center}
\small
\begin{tabular}{l|cccc}
\Xhline{2\arrayrulewidth}
\textbf{Training \textbackslash Test} & \multicolumn{2}{c}{ETH-XGaze}& \multicolumn{2}{c}{EYEDIAP} \\ 
& Train & Test  &  CS & FT \\ 
\hline
MPIIFaceGaze~\cite{zhang19_pami}  &  32.5 & 33.0 &  14.3 & 24.5 \\
ETH-XGaze Train~\cite{Zhang2020ETHXGaze} & - & - & 8.8 & 13.0 \\ 
Gaze360~\cite{gaze360_2019}  & 17.5 & 18.2 &  7.6  &  \textbf{12.6}\\
\hline
MPII-NV-STED~\cite{Zheng2020NeurIPS} & 27.2 & 29.1  &  8.6 & 20.5  \\
\mpiitarget-128$^{\dagger}$~ &  13.0  & 14.1 & 6.3  & 15.7  \\ 
\hline
\mpiitarget~$^{\dagger}$ &  14.0  & 15.5 & 6.6  &  17.5 \\ 
\mpiitarget~(Mask)$^{\dagger}$ & \textbf{12.7} & \textbf{13.8} &  \textbf{5.6} &  16.4  \\ 
\Xhline{2\arrayrulewidth}
\end{tabular}
\end{center}
\caption{ Comparison of gaze estimation errors in degree. Each row corresponds to a training dataset, and the columns show the mean angular errors for each test dataset. $\dagger$ indicates our approach.}
\label{table:results}
\end{table}


The results are summarized in Table~\ref{table:results}. 
The upper block are real datasets, and the last four rows are extended synthetic datasets.
All are baseline model performance except the last row being trained with the proposed mask-guided model.
The middle block corresponds to the $128 \times 128$ models for comparison with ST-ED~\cite{Zheng2020NeurIPS}.

MPIIFaceGaze has the narrowest gaze range and resulted in the highest errors for all test datasets.
Our proposed synthetic dataset and model (last row) reduced these errors by $61\%$, $58\%$, $61\%$, and $33\%$ for each test dataset, respectively. 
ETH-XGaze and Gaze360 both contain a wider gaze range and perform better on other datasets but are still inferior to our synthetic data.
While MPII-NV-STED has a wide gaze range as our dataset, it does not effectively improve the performance.
This indicates the difficulty of maintaining ground-truth gaze labels through neural rendering, while our method faithfully reproduces the authentic gaze direction by sampling the original appearance.
As a result, our method achieved the best performance on ETH-XGaze and EYEDIAP CS.
Contrary to earlier reports~\cite{Zhang2020ETHXGaze}, the lower resolution model (MPII-NV-128) resulted in slightly better performance in our setting.

The only exception was the EYEDIAP FT subset, where better performance was obtained when using real data.
EYEDIAP FT has a larger offset between gaze and head pose due to the use of physical gaze targets, and our data synthesized based on head pose cannot fully reproduce the target gaze distribution (Fig.~\ref{fig:distribution}). 
For further analysis, refer to the supplementary material.

\paragraph{Further Comparison with ST-ED~\cite{Zheng2020NeurIPS}}
As shown in Fig.~\ref{fig:sted-samples}, ST-ED cannot preserve the identity of MPIIFaceGaze because its model was pre-trained on the GazeCapture dataset.
Thus, we further compare our approach with ST-ED by using GazeCapture as the source dataset.

We randomly chose 1,000 out of the 1,374 subjects in GazeCapture and further randomly chose 30 images from each subject.
We used the $128 \times 128$ image resolution and sampled 12 new head poses from ETH-XGaze for each source image. 
The gaze estimation error of the baseline model on the ETH-XGaze Train set was \textbf{20.6} and \textbf{26.3} degrees for our synthetic data and ST-ED, respectively.
This again proved that the neural rendering approach cannot yet provide accurate image and gaze label pairs for training.

\paragraph{Effect of Head Pose Prior}
As discussed earlier, the head pose distribution of the target dataset can be obtained from unlabelled images.
However, in practice, there may be use cases where target environment samples are totally unknown.
To represent such cases, we further evaluate the performance by synthesizing samples without relying on any prior knowledge about the target dataset.
Specifically, we assumed a zero mean normal distribution for both the yaw and the pitch of the gaze angle, and varied the standard deviation $\sigma$ from 5 to 40 degrees.
Source data is MPIIFaceGaze, and we tested on ETH-XGaze Train set and EYEDIAP CS.
We set the same random background augmentation, so that these datasets only differ in the head pose.

\begin{figure}[t]
\begin{center}
  \includegraphics[width=0.85\linewidth]{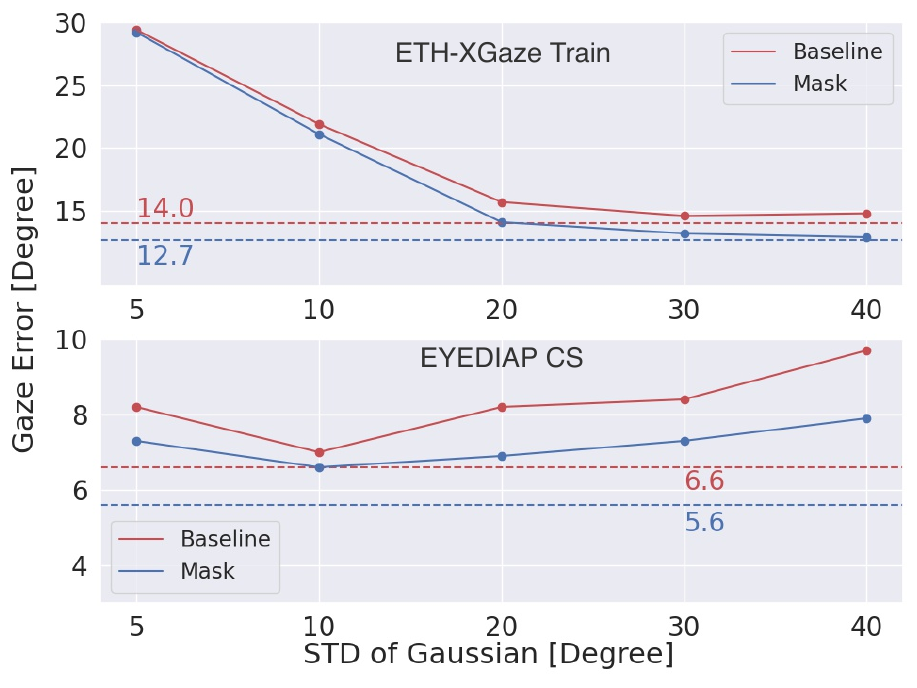}
\end{center}
  \caption {The gaze estimation errors of both baseline and mask-guided models with respect to the variance $\sigma$ of Gaussian sampling. The horizontal dashed lines correspond to the error reported in Table~\ref{table:results} directly using the target head pose distribution.}
\label{fig:gaussian-error-vs-sigma}
\end{figure}

Figure~\ref{fig:gaussian-error-vs-sigma} shows the gaze estimation errors with respect to $\sigma$ for ETH-XGaze Train set and EYEDIAP CS.
When $\sigma$ is very small, the synthetic dataset still cannot cover the gaze distributions in both cases, so the models do not perform well.
As $\sigma$ increases, the head pose coverage also increases and the performance approaches the best case scenario.
However, since EYEDIAP CS has a narrow gaze range (Fig.~\ref{fig:distribution}) compared to ETH-XGaze, the errors start to increase after $\sigma=20$. 
This indicates that, although synthesizing data over an excessively wide range may adversely affect the performance, sufficient performance can be obtained without prior knowledge on head pose distribution.

\subsection{Ablation Studies}\label{ablation-study}

We evaluate the effect of data augmentation and mask-guided model using \mpiitarget~(tested on the ETH-XGaze Train set).
We also use the frontal camera of the ETH-XGaze Train set as another source dataset to see the upper bound performance of our approach on the ETH-XGaze Test set.
From the frontal image, we synthesize images under head poses corresponding to all 18 cameras (XGazeF-NV).
In this within-dataset setting, the best performance using the real ETH-XGaze Train set is \textbf{4.5} degrees.
In the first four rows of Table~\ref{table:ablation-study}, we can observe the performance gain by adding random colors, random scene images, and weak lighting.
Black-only background tends to overfit and is effectively alleviated by adding random colors and random scenes.
We keep $40\%$ color background images to avoid poor generalization on simple background test data.
Finally, the increased diversity of lighting made the model more robust.
For the data augmentation effect on the mask-guided model, refer to the supplementary material.

In addition, we compare the proposed mask-guided model with some existing domain adaptation methods in the last four rows in Table~\ref{table:ablation-study}.
We use our implementations of SimGAN~\cite{Shrivastava_2017_CVPR}, DANN~\cite{ganin2015unsupervised}, and PADACO~\cite{9009467}, all using the architecture of the baseline gaze estimation network.
These implementation details can be found in the supplementary material.
Overall, these domain adaptation methods cannot consistently outperform the baseline model (fourth row).
In contrast, our mask-guided model effectively reduced the error by benefiting from the synthesis process.
The best performance (8.3 degrees) using XGazeF-NV is comparative with the result using the real ETH-XGaze Train set, while indicating the effect of remaining domain gaps.


\begin{table}[t]
\small
\begin{center}
\begin{tabular}{l|ccc}
\Xhline{2\arrayrulewidth}
\textbf{Ablations\textbackslash Datasets}  & MPII-NV & XGazeF-NV   \\
\hline
Black & 26.0  & 21.6 \\
+ Color (1:1) & 17.8 &  18.7  \\
+ Scene (1:1:3) & 14.4  & 12.9  \\
+ Weak-light & 14.0  &  11.2\\
\hline
SimGAN~\cite{Shrivastava_2017_CVPR} & 14.2& 10.0 \\
DANN~\cite{ganin2015unsupervised}\ & 13.6  & 19.1\\
PADACO~\cite{9009467} & 13.2 & 28.7\\
Mask-guided (ours)& \textbf{12.7} & \textbf{8.3} \\
\Xhline{2\arrayrulewidth}
\end{tabular}
\end{center}
\caption{Ablation study for analyzing the data augmentation and models. The data augmentation components are evaluated on the baseline model. 
}
\label{table:ablation-study}
\end{table}

\subsection{Comparison of Reconstruction Methods}

\begin{figure}[t]
\begin{center}
  \includegraphics[width=0.8\linewidth]{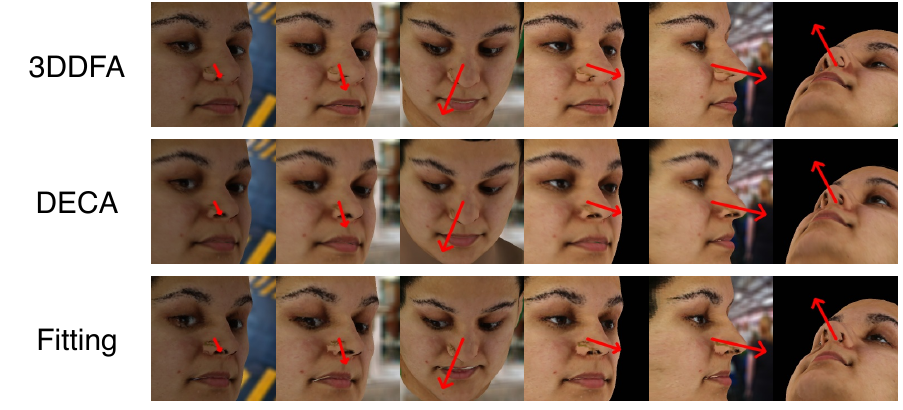}
\end{center}
   \caption{Examples of the synthesized XGazeF-NV datasets using the three different reconstruction methods.}
\label{fig:samples-3-recons}
\end{figure}

We further analyzed the influence of different reconstruction methods on the gaze estimation errors.
We employed DECA~\cite{Yao2021DECA}, which reaches the state-of-the-art mean shape reconstruction error on NoW benchmark~\cite{RingNetCVPR2019}.
While 3DDFA and DECA are both learning-based, they are trained with different 3D models: BFM~\cite{bfm09} and FLAME~\cite{FLAME:SiggraphAsia2017}, respectively.
As another baseline named 3DMM-Fitting, we simply fit the BFM model~\cite{bfm09} to the detected 68 2D facial landmarks.
Although the output formats of these methods are different, we manually converted and aligned them to meet the underlying assumption of our projective matching procedure.
We synthesized three versions of XGazeF-NV simultaneously, under the same random augmentation conditions, as shown in Fig.~\ref{fig:samples-3-recons}.

We used the baseline model and tested it on the ETH-XGaze Test set. 
The gaze estimation errors are \textbf{11.83} (3DDFA), \textbf{11.29} (DECA), and \textbf{11.79} (3DMM-Fitting), respectively.
We can observe that the influence of the reconstruction accuracy is relatively minor compared to other factors and even the simplest baseline works sufficiently well.

\section{Conclusion}

In this work, we presented a novel learning-by-synthesis pipeline for appearance-based full-face gaze estimation.
Our approach utilizes 3D face reconstruction to synthesize training datasets with novel head poses, while keeping accurate gaze labels via projective matching. 
We also proposed the mask-guided gaze estimation model with synthetic data augmentation.
Through experiments, our approach effectively improved the model and achieved better performance than the state-of-the-art neural rendering approach.

As discussed in the experiment, it is still difficult for our method to extend the limited head-gaze offset distribution in the source dataset.
It is important future work to explore learning-by-synthesis approaches to cover different data diversity requirements.
All datasets used in this work were collected with the approval of the IRB or the consent of the participants~\cite{zhang2017s,Zhang2020ETHXGaze,FunesMora_ETRA_2014,7780608,gaze360_2019}.
Although the proposed method creates synthetic faces, ethical issues are minimal because the method cannot extend the diversity of human faces by synthesizing new identities.

\section*{Acknowledgement}
This work was supported by JSPS KAKENHI Grant Number JP21K11932.

{\small
\bibliographystyle{ieee_fullname}
\bibliography{main}
}

\end{document}